# DYNAMICS OF THE ORTHOGLIDE PARALLEL ROBOT

Damien CHABLAT[1], Philippe WENGER[2], Stefan STAICU[3]

*Articolul stabileşte relaţii matriceale pentru cinematica şi dinamica robotului paralel Orthoglide prevăzut cu trei acţionori prismatici concurenţi. Aceştia sunt aranjaţi în raport cu sistemul cartezian de coordonate astfel încât direcţiile lor să fie normale unele faţă de celelalte. Trei lanţuri cinematice identice, conectate la platforma mobilă, sunt localizate în trei plane perpendiculare unul pe celălalt. Cunoscând poziţia şi mişcarea de translaţie a platformei, se dezvoltă problema de cinematică inversă şi se determină poziţia, viteza şi acceleraţia fiecărui element al robotului. În continuare, principiul lucrului mecanic virtual este folosit în problema de dinamică inversă. Câteva ecuaţii matriceale oferă expresii recurente şi grafice pentru forţele active şi puterile mecanice ale celor trei acţionori.*

*Recursive matrix relations for kinematics and dynamics of the Orthoglide parallel robot having three concurrent prismatic actuators are established in this paper. These are arranged according to the Cartesian coordinate system with fixed orientation, which means that the actuating directions are normal to each other. Three identical legs connecting to the moving platform are located on three planes being perpendicular to each other too. Knowing the position and the translation motion of the platform, we develop the inverse kinematics problem and determine the position, velocity and acceleration of each element of the robot. Further, the principle of virtual work is used in the inverse dynamic problem. Some matrix equations offer iterative expressions and graphs for the input forces and the powers of the three actuators.*

**List of symbols**

$a_{k,k-1}$ : orthogonal relative transformation matrix

$\vec{u}_1, \vec{u}_2, \vec{u}_3$ : three orthogonal unit vectors

$\alpha$ : orientation angle of the slider about the guide-way

$\varphi_{k,k-1}$ : relative rotation angle of $T_k$ rigid body

$\vec{\omega}_{k,k-1}$ : relative angular velocity of $T_k$

$\vec{\omega}_{k0}$ : absolute angular velocity of $T_k$

---
[1] Prof., Institut de Recherche en Communications et Cybernétique de Nantes, France
[2] Prof., Institut de Recherche en Communications et Cybernétique de Nantes, France
[3] Prof., Département de Mécanique, Université Polytechnique de Bucarest, Roumanie, e-mail :stefanstaicu@yahoo.com

$\tilde{\omega}_{k,k-1}$ : skew symmetric matrix associated to the angular velocity $\vec{\omega}_{k,k-1}$

$\vec{\varepsilon}_{k,k-1}$ :   relative angular acceleration of $T_k$

$\tilde{\varepsilon}_{k0}$ :    absolute angular acceleration of $T_k$

$\tilde{\varepsilon}_{k,k-1}$ : skew symmetric matrix associated to the angular acceleration $\vec{\varepsilon}_{k,k-1}$

$\vec{r}^{A}_{k,k-1}$ : relative position vector of the centre of $A_k$ joint

$\vec{v}^{A}_{k,k-1}$ : relative velocity of the centre $A_k$

$\vec{\gamma}^{A}_{k,k-1}$ : relative acceleration of the centre $A_k$

$m_k$ :   mass of $T_k$ rigid body

$\hat{J}_k$ :    symmetric matrix of tensor of inertia of $T_k$ about the link-frame $A_k x_k y_k z_k$

$J_1, J_2$ : two Jacobian matrices of the manipulator

$f^{A}_{10}, f^{B}_{10}, f^{C}_{10}$ : forces of three actuators pointing about the $A_1 z^{A}_1$, $B_1 z^{B}_1$, $C_1 z^{C}_1$ axes

## 1. Introduction

Generally, the mechanism of a parallel robot has two platforms: one of them is attached to the fixed reference frame and the other one can have arbitrary motions in its workspace. Some movable legs, made up as serial robots, connect the moving platform to the fixed platform. Typically, a parallel mechanism is said to be *symmetrical* if it satisfies the following conditions: the number of legs is equal to the number of degrees of freedom of the moving platform, one actuator, which can be mounted at or near the fixed base, controls every limb and the location and the number of actuated joints in all the limbs are the same (Tsai [1]).

For two decades, parallel manipulators attracted to the attention of more and more researches that consider them as valuable alternative design for robotic mechanisms [2], [3], [4]. As stated by a number of authors [1], conventional serial kinematical machines have already reached their dynamic performance limits, which are bounded by high stiffness of the machine components required to support sequential joints, links and actuators.

The parallel robots are spatial mechanisms with supplementary characteristics, compared with the serial architecture manipulators such as: more rigid structure, important dynamic charge capacity, high orientation accuracy, stabile functioning as well as good control of velocity and acceleration limits. However, most existing parallel manipulators have limited and complicated workspace with singularities and highly non-isotropic input-output relations [5].

Research in the field of parallel manipulators began with the most known application in the flight simulator with six degrees of freedom, which is in fact the Stewart-Gough platform (Stewart [6]; Merlet [7]; Parenti-Castelli and Di Gregorio [8]). The Star parallel manipulator (Hervé and Sparacino [9]) and the Delta parallel robot (Clavel [10]; Tsai and Stamper [11]; Staicu [12]) equipped with three motors, which have a parallel setting, train on the effectors in a three-degrees-of-freedom general translation motion.

While the kinematics has been studied extensively during the last two decades, fewer papers can be focused on the dynamics of parallel robots. When good dynamic performance and precise positioning under high load are required, the dynamic model is important for their control. The analysis of parallel robots is usually implemented trough analytical methods in classical mechanics [13], in which projection and resolution of equations on the reference axes are written in a considerable number of cumbersome, scalar relations and the solutions are rendered by large scale computation together with time consuming computer codes. Geng [14] developed Lagrange's equations of motion under some simplifying assumptions regarding the geometry and inertia distribution of the manipulator. Dasgupta and Mruthyunjaya [15] used the Newton-Euler approach to develop closed-form dynamic equations of Stewart platform, considering all dynamic and gravity effects as well as viscous friction at joints. However, to the best of our knowledge, these are no efficient dynamic modelling approach available for parallel manipulators. In recent years, several new kinematical structures have been proposed that possess higher isotropy [16], [17], [18], [19], [20].

The objective of this paper is to analyse the kinematics and dynamics of the Orthoglide parallel robot, which is well adapted to the applications of precision assembly machines. In design, the three actuators are arranged according to the Cartesian coordinate space, which means that the actuating directions are normal to each other and the joints connecting to the moving platform are located on three planes being perpendicular to each other too. Proposed by Wenger and Chablat [21], [22], the prototype of the manipulator has good kinetostatic performance and some technological advantages such as: symmetrical design, regular workspace shape properties with a bounded velocity amplification factor and low inertia effects.

In the present paper we focus our attention on a recursive matrix method, which is adopted to derive the kinematics model and the inverse dynamics equations of the spatial Orthoglide parallel robot [23], which has three translation degrees of freedom (Fig. 1).

## 2. Inverse kinematics

The mechanism input of the manipulator is made up of three actuated orthogonal prismatic joints. The output body is connected to the prismatic joints through a set of three identical kinematical chains.

The architecture of one of the three parallel closed chains of the Orthoglide manipulator is formally described as $PRP_aR$ mechanism, where $P, R$ and $P_a$ denote the prismatic, revolute and parallelogram joints, respectively. So, the topological structure consists in an active prismatic system, a passive revolute joint, an intermediate mechanism with four revolute links that connect four bars, which are parallel two by two, ending with a passive revolute link connected to the moving platform. Inside each chain, the parallelogram mechanism is used and oriented in a manner that the end-effector is restricted to *translation* movement only. The arrangement of the joints in the chains has been defined to eliminate any constraint singularity in the Cartesian workspace [22], [23], [24].

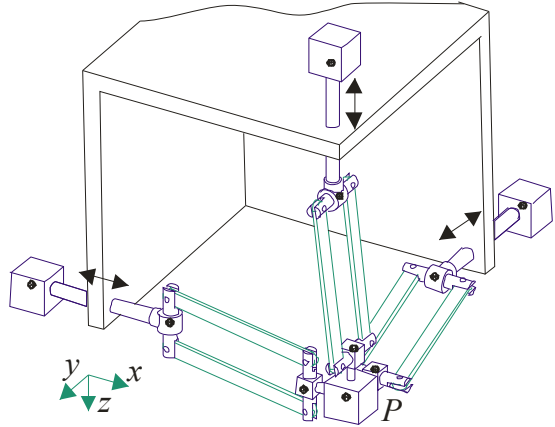

Fig. 1 Orthoglide parallel robot

Let us locate a fixed reference frame $Ox_0y_0z_0(T_0)$ at the intersection point of three axes of actuated prismatic joints, about which the three-degrees-of-freedom manipulator moves. It has three legs of known dimensions and masses. To simplify the graphical image of the kinematical scheme of the mechanism, in the follows we will represent the intermediate reference systems by only two axes, so as is used in most of books [1], [4], [5], [7]. The $z_k$ axis is represented, of course, for each component element $T_k$. We mention that the relative rotation or relative translation with $\varphi_{k,k-1}$ angle or $\lambda_{k,k-1}$ displacement of $T_k$ body most be always pointing about or along the direction $z_k$.

The first element of leg $A$ is one of the three *sliders* of the robot. It is a homogenous rod of length $A_1A_2 = l_1$ and mass $m_1$, moving horizontally along the fixed $A_1z_1^A$ axis with a displacement $\lambda_{10}^A$. The centre of the transmission rod $A_3A_6 = l_2$ is denoted as $A_2$. This link is connected to the frame $A_2x_2^Ay_2^Az_2^A$ (called $T_2^A$) and it has a relative rotation with the angle $\varphi_{21}^A$, so that $\omega_{21}^A = \dot{\varphi}_{21}^A$ and $\varepsilon_{21}^A = \ddot{\varphi}_{21}^A$. It has the mass $m_2$ and the central tensor of inertia $\hat{J}_2$. Further one, two identical and parallel bars $A_3 A_4$ and $A_6 A_7$ with same length $l_3$ rotate about the $T_2^A$ frame with the angle $\varphi_{32}^A = \varphi_{62}^A$. They have also the same mass $m_3$ and the same tensor of inertia $\hat{J}_3$. The four-bar parallelogram is closed by an element $T_4^A$, which is identical with $T_2^A$. Its tensor of inertia is $\hat{J}_4$. This element rotates with the relative angle $\varphi_{43}^A = \varphi_{32}^A$ (Fig. 2).

The centre $A_5$ of the interval between the two revolute joints connects the moving platform $A_5 x_5^A y_5^A z_5^A (T_5^A)$. The platform of the robot may be a cube of masse $m_5$, central tensor of inertia $\hat{J}_G$ and side dimension $l$, which rotate relatively by an angle $\varphi_{54}^A$ with respect to the neighbouring body $T_4^A$. Finally, another reference system $G x_G y_G z_G$ is located at the centre $G$ of the cubic moving platform.

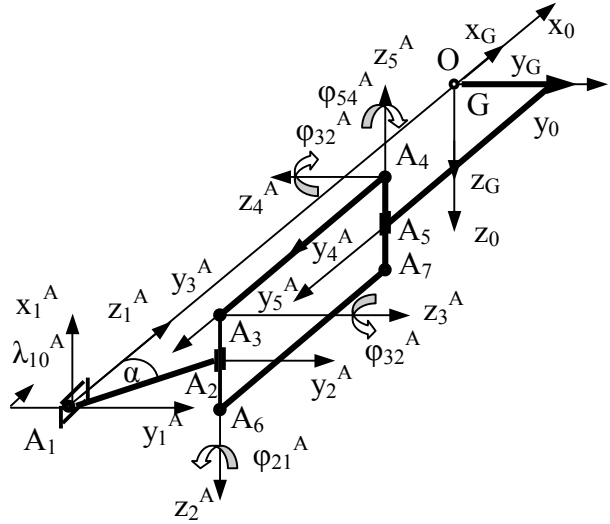

Fig. 2 Kinematical scheme of first leg $A$ of the mechanism

Due to the special arrangement of the four-bar parallelograms and the three prismatic joints at points $A_1, B_1, C_1$, the mechanism has three translation degrees of freedom. This unique characteristic is useful in many applications, such as a $x-y-z$ positioning device.

At the central configuration, we consider that all legs are initially extended at equal length and that the angles giving the orientation of the three sliders about their guide-ways are $\alpha_A = \alpha_B = \alpha_C = \alpha$.

In the followings, we apply the method of successive displacements to geometric analysis of closed-loop chains and we note that a joint variable is the displacement required to move a link from the initial location to the actual position. If every link is connected to least two other links, the chain forms one or more independent closed-loops. The variable angles $\varphi_{k,k-1}$ of rotation about the joint axes $z_k$ are the parameters needed to bring the next link from a reference configuration to the next configuration. We call the matrix $a^{\varphi}_{k,k-1}$, for example, the orthogonal transformation $3 \times 3$ matrix of relative rotation with the angle $\varphi^A_{k,k-1}$ of link $T^A_k$ around $z^A_k$ axis.

In the study of the kinematics of robot manipulators, we are interested in deriving a matrix equation relating the location of an arbitrary $T_k$ body to the joint variables. When the change of coordinates is successively considered, the corresponding matrices are multiplied.

So, starting from the reference origin $O$ and pursuing the three legs $A$, $B$, $C$ we obtains the following transformation matrices [25]

$$a_{10} = a_1, \ a_{21} = a^{\varphi}_{21}a_2, \ a_{32} = a^{\varphi}_{32}a_3$$
$$a_{43} = a^{\varphi}_{32}a_4, \ a_{54} = a^{\varphi}_{54}a_2, \ a_{62} = a_{32}$$
$$b_{10} = a_5, \ b_{21} = b^{\varphi}_{21}a_2, \ b_{32} = b^{\varphi}_{32}a_3 \quad (1)$$
$$b_{43} = b^{\varphi}_{32}a_4, \ b_{54} = b^{\varphi}_{54}a_2, \ b_{62} = b_{32}$$
$$c_{10} = a_6, \ c_{21} = c^{\varphi}_{21}a_2, \ c_{32} = c^{\varphi}_{32}a_3$$
$$c_{43} = c^{\varphi}_{32}a_4, \ c_{54} = c^{\varphi}_{54}a_2, \ c_{62} = c_{32}$$

where we denoted

$$a_1 = \begin{bmatrix} 0 & 0 & -1 \\ 0 & 1 & 0 \\ 1 & 0 & 0 \end{bmatrix}, a_2 = \begin{bmatrix} 0 & 0 & 1 \\ 0 & 1 & 0 \\ -1 & 0 & 0 \end{bmatrix}, a_3 = \begin{bmatrix} 0 & 0 & -1 \\ -1 & 0 & 0 \\ 0 & 1 & 0 \end{bmatrix}$$

$$a_4 = \begin{bmatrix} -1 & 0 & 0 \\ 0 & 1 & 0 \\ 0 & 0 & -1 \end{bmatrix}, a_5 = \begin{bmatrix} -1 & 0 & 0 \\ 0 & 0 & 1 \\ 0 & 1 & 0 \end{bmatrix}, a_6 = \begin{bmatrix} 0 & -1 & 0 \\ 1 & 0 & 0 \\ 0 & 0 & 1 \end{bmatrix}$$

$$a^{\varphi}_{k,k-1} = \begin{bmatrix} \cos\varphi^A_{k,k-1} & \sin\varphi^A_{k,k-1} & 0 \\ -\sin\varphi^A_{k,k-1} & \cos\varphi^A_{k,k-1} & 0 \\ 0 & 0 & 1 \end{bmatrix}, \ a_{k0} = \prod_{j=1}^{k} a_{k-j+1,k-j} \quad (k=1,2,...,5). \quad (2)$$

The translation conditions for the platform are given by the following identities

$$a^{\circ T}_{50} a_{50} = b^{\circ T}_{50} b_{50} = c^{\circ T}_{50} c_{50} = I, \quad (3)$$

where

$$a_{50}^\circ = \begin{bmatrix} 0 & -1 & 0 \\ -1 & 0 & 0 \\ 0 & 0 & -1 \end{bmatrix}, b_{50}^\circ = \begin{bmatrix} 0 & 0 & -1 \\ 0 & -1 & 0 \\ -1 & 0 & 0 \end{bmatrix}, c_{50}^\circ = \begin{bmatrix} -1 & 0 & 0 \\ 0 & -1 & 0 \\ 0 & 0 & 1 \end{bmatrix}. \quad (4)$$

From these relations, one obtains the following relations between angles
$$\varphi_{54}^A = \varphi_{21}^A, \varphi_{54}^B = \varphi_{21}^B, \varphi_{54}^C = \varphi_{21}^C. \quad (5)$$

The three concurrent displacements $\lambda_{10}^A, \lambda_{10}^B, \lambda_{10}^C$ of the actuators $A_1, B_1, C_1$ are the joint variables that give the input vector $\vec{\lambda}_{10}$ of the instantaneous position of the mechanism. But, the objective of the inverse geometric problem is to find the input vector $\vec{\lambda}_{10}$ and the position of the robot with the given three absolute coordinates of the center $G$ of the platform: $x_0^G, y_0^G, z_0^G$.

Supposing, for example, that the known motion of the mass center $G$ of the platform is expressed by the following relations
$$\vec{r}_0^G = [x_0^G \quad y_0^G \quad z_0^G]$$
$$x_0^G = x_0^{G*}(1 - \cos\frac{\pi}{3}t), y_0^G = y_0^{G*}(1 - \cos\frac{\pi}{3}t), z_0^G = z_0^{G*}(1 - \cos\frac{\pi}{3}t), \quad (6)$$

the inputs $\lambda_{10}^A, \lambda_{10}^B, \lambda_{10}^C$ of the manipulators and the variables $\varphi_{21}^A, \varphi_{32}^A, \varphi_{21}^B, \varphi_{32}^B, \varphi_{21}^C, \varphi_{32}^C$ will be given by the following geometrical conditions

$$\vec{r}_{10}^A + \sum_{k=1}^{4} a_{k0}^T \vec{r}_{k+1,k}^A + a_{50}^T \vec{r}_5^{GA} = \vec{r}_{10}^B + \sum_{k=1}^{4} b_{k0}^T \vec{r}_{k+1,k}^B + b_{50}^T \vec{r}_5^{GB} = \vec{r}_{10}^C + \sum_{k=1}^{4} c_{k0}^T \vec{r}_{k+1,k}^C + c_{50}^T \vec{r}_5^{GC} = \vec{r}_0^G, \quad (7)$$

where, for example, one denoted
$$\vec{u}_1 = \begin{bmatrix} 1 \\ 0 \\ 0 \end{bmatrix}, \vec{u}_2 = \begin{bmatrix} 0 \\ 1 \\ 0 \end{bmatrix}, \vec{u}_3 = \begin{bmatrix} 0 \\ 0 \\ 1 \end{bmatrix}, \tilde{u}_3 = \begin{bmatrix} 0 & -1 & 0 \\ 1 & 0 & 0 \\ 0 & 0 & 0 \end{bmatrix}$$

$$\vec{r}_{10}^A = (\lambda_{10}^A - l_1 \cos\alpha - l_3 - \frac{l}{2})a_{10}^T \vec{u}_3$$

$$\vec{r}_{21}^A = [0 \quad l_1 \sin\alpha \quad l_1 \cos\alpha]^T, \quad \vec{r}_{32}^A = -\frac{l_2}{2}\vec{u}_3 \quad (8)$$

$$\vec{r}_{43}^A = -l_3\vec{u}_2, \vec{r}_{54}^A = \frac{l_2}{2}\vec{u}_1, \vec{r}_5^{GA} = [l_1 \sin\alpha \quad -\frac{l}{2} \quad 0]^T.$$

Actually, these equations means that there is the inverse geometric solution for the manipulator, given through following analytical relations

$$\sin\varphi_{32}^A = -\frac{z_0^G}{l_3}, \quad \sin\varphi_{21}^A = \frac{y_0^G}{l_3 \cos\varphi_{32}^A}, \quad \lambda_{10}^A = x_0^G + l_3(1 - \cos\varphi_{21}^A \cos\varphi_{32}^A)$$

$$\sin\varphi_{32}^B = -\frac{x_0^G}{l_3}, \quad \sin\varphi_{21}^B = \frac{z_0^G}{l_3 \cos\varphi_{32}^B}, \quad \lambda_{10}^B = y_0^G + l_3(1 - \cos\varphi_{21}^B \cos\varphi_{32}^B) \quad (9)$$

$$\sin\varphi^C_{32} = -\frac{y^G_0}{l_3}, \quad \sin\varphi^C_{21} = \frac{x^G_0}{l_3 \cos\varphi^C_{32}}, \quad \lambda^C_{10} = z^G_0 + l_3(1 - \cos\varphi^C_{21}\cos\varphi^C_{32}).$$

In that follows, we determine, the velocities and the accelerations of the robot, supposing that the translation motion of the moving platform is known.

The motions of the component elements of each leg (for example the leg *A*) are characterized by the following skew symmetric matrices [26]

$$\tilde{\omega}^A_{k0} = a_{k,k-1}\tilde{\omega}^A_{k-1,0}a^T_{k,k-1} + \omega^A_{k,k-1}\tilde{u}_3, \quad (k=2,\dots,5), \tag{10}$$

which are *associated* to the absolute angular velocities given by the recurrence relations

$$\vec{\omega}^A_{k0} = a_{k,k-1}\vec{\omega}^A_{k-1,0} + \omega^A_{k,k-1}\vec{u}_3, \quad \omega^A_{k,k-1} = \dot{\varphi}^A_{k,k-1}. \tag{11}$$

Following relations give the velocities $\vec{v}^A_{k0}$ of the joints $A_k$

$$\vec{v}^A_{k0} = a_{k,k-1}\left\{\vec{v}^A_{k-1,0} + \tilde{\omega}^A_{k-1,0}\vec{r}^A_{k,k-1}\right\}, \quad \vec{v}^A_{10} = \dot{\lambda}^A_{10}\vec{u}_3. \tag{12}$$

If the other two kinematical chains of the manipulator are pursued, analogous relations can be easily obtained.

Equations (3) and (7) can be differentiated with respect to time to obtain the following *matrix conditions of connectivity* [27]

$$\omega^A_{21}\vec{u}^T_i a^T_{20}\vec{u}_3 + \omega^A_{54}\vec{u}^T_i a^T_{50}\vec{u}_3 = 0$$
$$v^A_{10}\vec{u}^T_i a^T_{10}\vec{u}_3 + l_3\omega^A_{21}\vec{u}^T_i a^T_{20}\vec{u}_3 a^T_{32}\vec{u}_2 + l_3\omega^A_{32}\vec{u}^T_i a^T_{30}\tilde{u}_3\vec{u}_2 = \vec{u}^T_i \dot{\vec{r}}^G_0, \quad (i=1,2,3), \tag{13}$$

where $\tilde{u}_1, \tilde{u}_2, \tilde{u}_3$ are skew-symmetric matrices associate to three orthogonal unit vectors $\vec{u}_1, \vec{u}_2, \vec{u}_3$. From these equations, relative velocities $v^A_{10}, \omega^A_{21}, \omega^A_{32}$ and $\omega^A_{54} = \omega^A_{21}$ result as functions of the translation velocity of the platform. The relations (13) give the *complete* Jacobian matrix of the manipulator. This matrix is a fundamental element for the analysis of the robot workspace and the particular configurations of singularities where the manipulator becomes uncontrollable.

Rearranging, above nine constraint equations (9) of the Orthoglide robot can immediately written as follows

$$z^{G2}_0 + y^{G2}_0 + (x^G_0 + l_3 - \lambda^A_{10})^2 = l^2_3$$
$$x^{G2}_0 + z^{G2}_0 + (y^G_0 + l_3 - \lambda^B_{10})^2 = l^2_3 \tag{14}$$
$$y^{G2}_0 + x^{G2}_0 + (z^G_0 + l_3 - \lambda^C_{10})^2 = l^2_3,$$

where the "zero" position $\vec{r}^{0G}_0 = [0\ \ 0\ \ 0]$ corresponds to the joints variables $\vec{\lambda}^0_{10} = [0\ \ 0\ \ 0]$. The derivative with respect to time of conditions (14) leads to the matrix equation

$$J_1\dot{\vec{\lambda}}_{10} = J_2\dot{\vec{r}}^G_0. \tag{15}$$

Matrices $J_1$ and $J_2$ are, respectively, the inverse and forward Jacobian of the manipulator and can be expressed as

$$J_1 = diag\{\alpha_1 \quad \alpha_2 \quad \alpha_3\}, J_2 = \begin{bmatrix} \alpha_1 & y_0^G & z_0^G \\ x_0^G & \alpha_2 & z_0^G \\ x_0^G & y_0^G & \alpha_3 \end{bmatrix}, \tag{16}$$

with

$$\alpha_1 = x_0^G + l_3 - \lambda_{10}^A;\ \alpha_2 = y_0^G + l_3 - \lambda_{10}^B;\ \alpha_3 = z_0^G + l_3 - \lambda_{10}^C. \tag{17}$$

The three kinds of singularities of the three closed-loop kinematical chains can be determined through the analysis of two Jacobian matrices $J_1$ and $J_2$.

Let us assume that the robot has a first virtual motion determined by the linear velocities $v_{10a}^{Av} = 1$, $v_{10a}^{Bv} = 0$, $v_{10a}^{Cv} = 0$. The characteristic virtual velocities are expressed as functions of the position of the mechanism by the general kinematical constraints equations (13). Other two sets of relations of connectivity can be obtained if one considers successively: $v_{10b}^{Bv} = 1, v_{10b}^{Cv} = 0$, $v_{10b}^{Av} = 0$ and $v_{10c}^{Cv} = 1, v_{10c}^{Av} = 0, v_{10c}^{Bv} = 0$.

As for the relative accelerations $\gamma_{10}^A$, $\varepsilon_{21}^A$, $\varepsilon_{32}^A$ and $\varepsilon_{54}^A = \varepsilon_{21}^A$ of the manipulator, the derivatives with respect to time of the relations (13) give other following conditions of connectivity [28]

$$\begin{aligned}
&\varepsilon_{21}^A \vec{u}_i^T a_{20}^T \vec{u}_3 + \varepsilon_{54}^A \vec{u}_i^T a_{50}^T \vec{u}_3 = 0 \\
&\gamma_{10}^A \vec{u}_i^T a_{10}^T \vec{u}_3 + l_3 \varepsilon_{21}^A \vec{u}_i^T a_{20}^T \widetilde{u}_3 a_{32}^T \vec{u}_2 + l_3 \varepsilon_{32}^A \vec{u}_i^T a_{30}^T \widetilde{u}_3 \vec{u}_2 = \vec{u}_i^T \ddot{\vec{r}}_0^G - \\
&- l_3 \omega_{21}^A \omega_{21}^A \vec{u}_i^T a_{20}^T \widetilde{u}_3 \widetilde{u}_3 a_{32}^T \vec{u}_2 - l_3 \omega_{32}^A \omega_{32}^A \vec{u}_i^T a_{30}^T \widetilde{u}_3 \widetilde{u}_3 \vec{u}_2 - \\
&- 2l_3 \omega_{21}^A \omega_{32}^A \vec{u}_i^T a_{20}^T \widetilde{u}_3 a_{32}^T \widetilde{u}_3 \vec{u}_2,\ (i=1,2,3).
\end{aligned} \tag{18}$$

The angular accelerations $\vec{\varepsilon}_{k0}^A$ and the accelerations $\vec{\gamma}_{k0}^A$ of joints are given by some relations, obtained by deriving the relations (10), (11) and (12):

$$\begin{aligned}
\vec{\varepsilon}_{k0}^A &= a_{k,k-1} \vec{\varepsilon}_{k-1,0}^A + \varepsilon_{k,k-1}^A \vec{u}_3 + \omega_{k,k-1}^A a_{k,k-1} \widetilde{\omega}_{k-1,0}^A a_{k,k-1}^T \vec{u}_3 \\
\widetilde{\omega}_{k0}^A \widetilde{\omega}_{k0}^A + \widetilde{\varepsilon}_{k0}^A &= a_{k,k-1} \left( \widetilde{\omega}_{k-1,0}^A \widetilde{\omega}_{k-1,0}^A + \widetilde{\varepsilon}_{k-1,0}^A \right) a_{k,k-1}^T + \omega_{k,k-1}^A \omega_{k,k-1}^A \widetilde{u}_3 \widetilde{u}_3 + \varepsilon_{k,k-1}^A \widetilde{u}_3 + \\
&\quad + 2\omega_{k,k-1}^A a_{k,k-1} \widetilde{\omega}_{k-1,0}^A a_{k,k-1}^T \widetilde{u}_3 \\
\vec{\gamma}_{k0}^A &= a_{k,k-1} \left[ \vec{\gamma}_{k-1,0}^A + \left( \widetilde{\omega}_{k-1,0}^A \widetilde{\omega}_{k-1,0}^A + \widetilde{\varepsilon}_{k-1,0}^A \right) \vec{r}_{k,k-1}^A \right],\ \vec{\gamma}_{10}^A = \ddot{\lambda}_{10}^A \vec{u}_3
\end{aligned} \tag{19}$$

The relations (13), (18) represent the *inverse kinematics model* of the Orthoglide parallel robot. As application let us consider a manipulator, which has the following characteristics

$$x_0^{G*} = 0.05\,m,\ y_0^{G*} = 0.10\,m,\ z_0^{G*} = -0.20\,m$$

$$l = 0.20\,m, l_1 = 0.15\,m, l_2 = 0.08\,m, l_3 = 0.85\,m, l_4 = l_2, \alpha = \frac{\pi}{4}, \Delta t = 2\,s$$

$$m_1 = 0.35\,kg, m_2 = 0.2\,kg,\ m_3 = 2.5\,kg, m_4 = m_2, m_5 = 15\,kg, m_6 = m_3.$$

A program which implements the suggested algorithm is developed in MATLAB to solve first the inverse kinematics of the Orthoglide parallel robot. For illustration, it is assumed that for a period of two second the platform starts at rest from a central configuration and is moving in a general translation. A numerical study of the robot kinematics is carried out by computation of the input displacements $\lambda_{10}^A$, $\lambda_{10}^B$, $\lambda_{10}^C$, for example, of three prismatic actuators (Fig. 3, 4, 5).

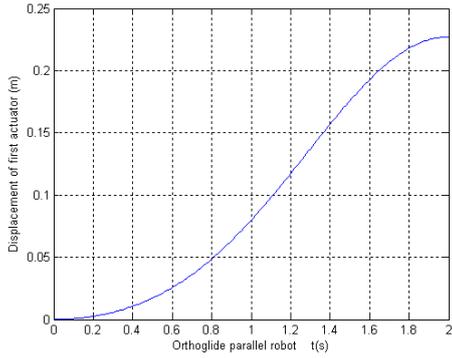
Fig. 3 Input displacement $\lambda_{10}^A$ of first actuator

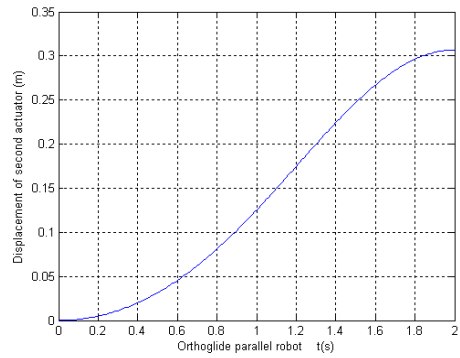
Fig. 4 Input displacement $\lambda_{10}^B$ of second actuator

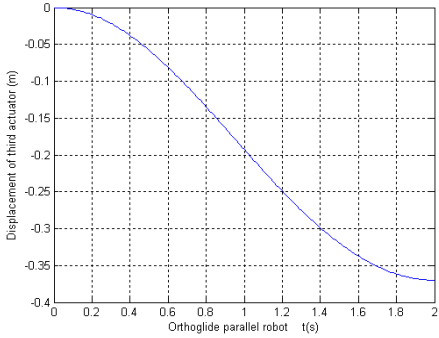
Fig. 5 Input displacement $\lambda_{10}^C$ of third actuator

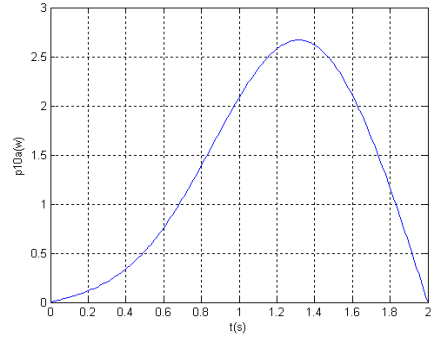
Fig. 6 Input power $p_{10}^A$ of first actuator

### 3. Inverse dynamics model

In the context of the real-time control, neglecting the frictions forces and considering the gravitational effects, the relevant objective of the dynamics is to determine the input forces, which must be exerted by the actuators in order to produce a given trajectory of the effector.

There are three methods, which can provide the same results concerning these actuating forces. The first one is using the Newton-Euler classic procedure [13], [15], [19], [29], the second one applies the Lagrange's equations and multipliers formalism [14], [30] and the third one is based on the principle of virtual work [1], [5], [25], [26]. In the inverse dynamic problem, in the present paper one applies the principle of virtual work in order to establish some recursive matrix relations for the forces and the powers of the three active systems.

Three input spatial concurrent forces $f_{10}^{j}$ and three powers $p_{10}^{j} = v_{10}^{j} f_{10}^{ji}$ ($j = A, B, C$) required in a given motion of the moving platform will easily be computed using a recursive procedure. Some independent pneumatic or hydraulic systems that generate three input forces $\vec{f}_{10}^{j} = f_{10}^{j} \vec{u}_3$, which are oriented along the axes $A_1 z_1^A$, $B_1 z_1^B$, $C_1 z_1^C$, control the motion of the three sliders of the robot.

Now, the parallel mechanism can artificially be transformed in a set of three open serial chains $C_j$ ($j = A, B, C$) subject to the constraints. This is possible by cutting successively the joints $A_5$, $B_5$, $C_5$ for the moving platform and $A_7$, $B_7$, $C_7$ for the four-bar parallelograms and taking their effects into account by introducing the corresponding constraint conditions. The first and more complicated open tree system includes the acting link and could comprise the moving platform.

The force of inertia of an arbitrary rigid body $T_k^A$, for example

$$\vec{f}_{k0}^{inA} = -m_k^A [\vec{\gamma}_{k0}^A + (\tilde{\omega}_{k0}^A \tilde{\omega}_{k0}^A + \tilde{\varepsilon}_{k0}^A) \vec{r}_k^{CA}] \tag{20}$$

and the resulting moment of the forces of inertia

$$\vec{m}_{k0}^{inA} = -[m_k^A \tilde{r}_k^{CA} \vec{\gamma}_{k0}^A + \hat{I}_k^A \vec{\varepsilon}_{k0}^A + \tilde{\omega}_{k0}^A \hat{I}_k^A \vec{\omega}_{k0}^A] \tag{21}$$

are determined with respect to the centre of its fist joint $A_k$. On the other hand, the wrench of two vectors $\vec{f}_k^{*A}$ and $\vec{m}_k^{*A}$ evaluates the influence of the action of the external and internal forces applied to the same element $T_k^A$ or of its weight $m_k^A \vec{g}$, for example:

$$\vec{f}_k^{*A} = 9.81 m_k^A a_{k0} \vec{u}_3, \ \vec{m}_k^{*A} = 9.81 m_k^A \tilde{r}_k^{CA} a_{k0} \vec{u}_3 \ (k = 1, 2, ..., 6). \tag{22}$$

Finally, two recursive relations generate the vectors

$$\begin{aligned}\vec{f}_k^A &= \vec{f}_{k0}^A + a_{k+1,k}^T \vec{f}_{k+1}^A \\ \vec{m}_k^A &= \vec{m}_{k0}^A + a_{k+1,k}^T \vec{m}_{k+1}^A + \tilde{r}_{k+1,k}^A a_{k+1,k}^T \vec{f}_{k+1}^A,\end{aligned} \tag{23}$$

where one denoted

$$\vec{f}_{k0}^A = -\vec{f}_{k0}^{inA} - \vec{f}_k^{*A}, \ \vec{m}_{k0}^A = -\vec{m}_{k0}^{inA} - \vec{m}_k^{*A}. \tag{24}$$

Considering three independent virtual motions of the robot, all virtual displacements and virtual velocities should be compatible with the virtual motions imposed by all kinematical constraints and joints at a given instant in time. By intermediate of the complete Jacobian matrix expressed by the conditions of connectivity (13), the absolute virtual velocities $\vec{v}_{k0}^v$, $\vec{\omega}_{k0}^v$ associated with all moving links are related to a set of independent *relative virtual velocities* $\vec{\omega}_{k,k-1}^v = \omega_{k,k-1}^v \vec{u}_3$.

Knowing the position and kinematics state of each link as well as the external forces acting on the robot, in that follow one apply the principle of virtual work for an inverse dynamic problem. The fundamental principle of the virtual work states that a mechanism is under dynamic equilibrium if and only if the total virtual work developed by all external, internal and inertia forces vanish during any general virtual displacement, which is compatible with the constraints imposed on the mechanism. Assuming that frictional forces at the joints are negligible, the virtual work produced by the forces of constraint at the joints is zero.

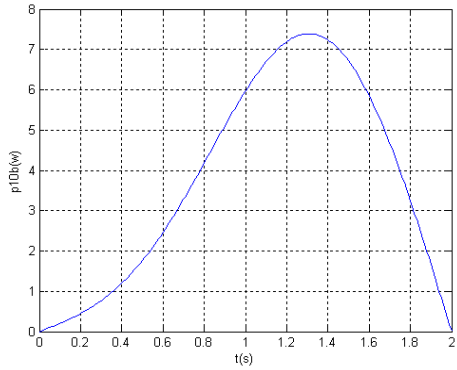 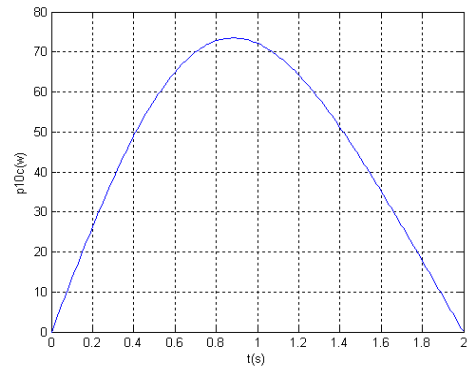

Fig. 7 Input power $p_{10}^B$ of second actuator    Fig. 8 Input power $p_{10}^C$ of third actuator

Applying *the fundamental equations of the parallel robots dynamics* established [31], following compact matrix relation results for the input force of first actuator

$$f_{10}^A = \vec{u}_3^T \left[ \vec{f}_1^A + \omega_{54a}^{Av} \vec{m}_5^A + \omega_{21a}^{Av} \vec{m}_2^A + \omega_{32a}^{Av}\left(\vec{m}_3^A + \vec{m}_4^A + \vec{m}_6^A\right) + \right. \\ \left. + \omega_{21a}^{Bv} \vec{m}_2^B + \omega_{32a}^{Bv}\left(\vec{m}_3^B + \vec{m}_4^B + \vec{m}_6^B\right) + \omega_{21a}^{Cv} \vec{m}_2^C + \omega_{32a}^{Cv}\left(\vec{m}_3^C + \vec{m}_4^C + \vec{m}_6^C\right) \right], \quad (25)$$

The relations (23), (25) represent the *inverse dynamics model* of the Orthoglide parallel robot.

Based on the algorithm derived from the above recursive relations, a computer program solve the inverse dynamics modelling of the robot, using the MATLAB software.

Assuming that the weights $m_k^A \vec{g}$ of compounding rigid bodies constitute the external forces acting on the robot during its evolution, a numerical computation in the dynamics is developed, based on the determination of the three active powers $p_{10}^A = v_{10}^A f_{10}^A$, $p_{10}^B = v_{10}^B f_{10}^B$, $p_{10}^C = v_{10}^C f_{10}^C$. The time-history evolution of the input powers $p_{10}^A$ (fig. 6), $p_{10}^B$ (fig. 7), $p_{10}^C$ (fig. 8) required by the actuators are plotted for a period of two second of platform's motion.

## 4. Conclusions

In the inverse kinematics analysis some exact relations that give in real-time the position, velocity and acceleration of each element of the parallel robot have been established in present paper. The dynamics model takes into consideration the masses and forces of inertia introduced by all component elements of the robot.

The new approach based on the principle of virtual work can eliminate all forces of internal joints and establishes a direct determination of the time-history evolution of forces and powers required by the actuators. The recursive matrix relations (25) represent the explicit equations of the dynamics simulation and can easily be transformed in a model for the automatic command of the Orthoglide parallel robot. Also, the method described above is quit available in forward and inverse mechanics of all serial or parallel mechanisms, the platform of which behaves in translation, rotation evolution or general 6-DOF motion.